\documentclass{article}

    \usepackage[preprint]{neurips_2024}

\usepackage[utf8]{inputenc} % allow utf-8 input
\usepackage[T1]{fontenc}    % use 8-bit T1 fonts
\usepackage{hyperref}       % hyperlinks
\usepackage{url}            % simple URL typesetting
\usepackage{booktabs}       % professional-quality tables
\usepackage{amsfonts}       % blackboard math symbols
\usepackage{nicefrac}       % compact symbols for 1/2, etc.
\usepackage{microtype}      % microtypography
\usepackage{xcolor}         % colors
\usepackage{graphicx}

\title{Exploring Vulnerabilities and Protections in Large Language Models: A Survey}

\author{%
  Frank Weizhen Liu\\
  Zscaler, Inc. \\
  \texttt{fliu@zscaler.com} \\
  \And
  Chenhui Hu \\
  Zscaler, Inc. \\
  \texttt{chenhuihu@zscaler.com} \\
}

\begin{document}

\maketitle

\begin{abstract}
As Large Language Models (LLMs) increasingly become key components in various AI applications, understanding their security vulnerabilities and the effectiveness of defense mechanisms is crucial. This survey examines the security challenges of LLMs, focusing on two main areas: Prompt Hacking and Adversarial Attacks, each with specific types of threats. Under Prompt Hacking, we explore Prompt Injection and Jailbreaking Attacks, discussing how they work, their potential impacts, and ways to mitigate them. Similarly, we analyze Adversarial Attacks, breaking them down into Data Poisoning Attacks and Backdoor Attacks. This structured examination helps us understand the relationships between these vulnerabilities and the defense strategies that can be implemented. The survey highlights these security challenges and discusses robust defensive frameworks to protect LLMs against these threats. By detailing these security issues, the survey contributes to the broader discussion on creating resilient AI systems that can resist sophisticated attacks.
\end{abstract}

\section{Introduction}

Since the release of ChatGPT~\cite{chatgpt2023}, large language models (LLMs) have garnered significant industrial and public attention. The surge in interest has prompted companies to either develop new products or integrate LLMs into existing ones~\cite{10336112}. However, LLMs are often trained on massive, uncurated datasets sourced from the internet~\cite{brown2020language}, which can contain sensitive information such as individual medical reports or government IDs \cite{kim2024propile}. This poses a risk of sensitive information leakage when LLMs are used in practice.

Moreover, while LLMs encapsulate a broad spectrum of human knowledge, they also have the potential to inadvertently teach users malicious skills, such as breaking into vehicles or synthesizing harmful substances. Despite the presence of safety controls in both open-source models like LLaMA2 and Gemma \cite{touvron2023llama,team2024gemma} and proprietary models such as GPT-4 and Claude-3 \cite{chatgpt2023,anthropic2024claude}, the dynamics of "shield and spear" persist, with attack strategies continuously evolving to become more sophisticated and potentially more destructive. This has given rise to a substantial community of AI skeptics and pessimists who highlight these risks \cite{ambartsoumean2023ai}.

\begin{figure}[ht]
    \centering
    \includegraphics[width=0.95\textwidth]{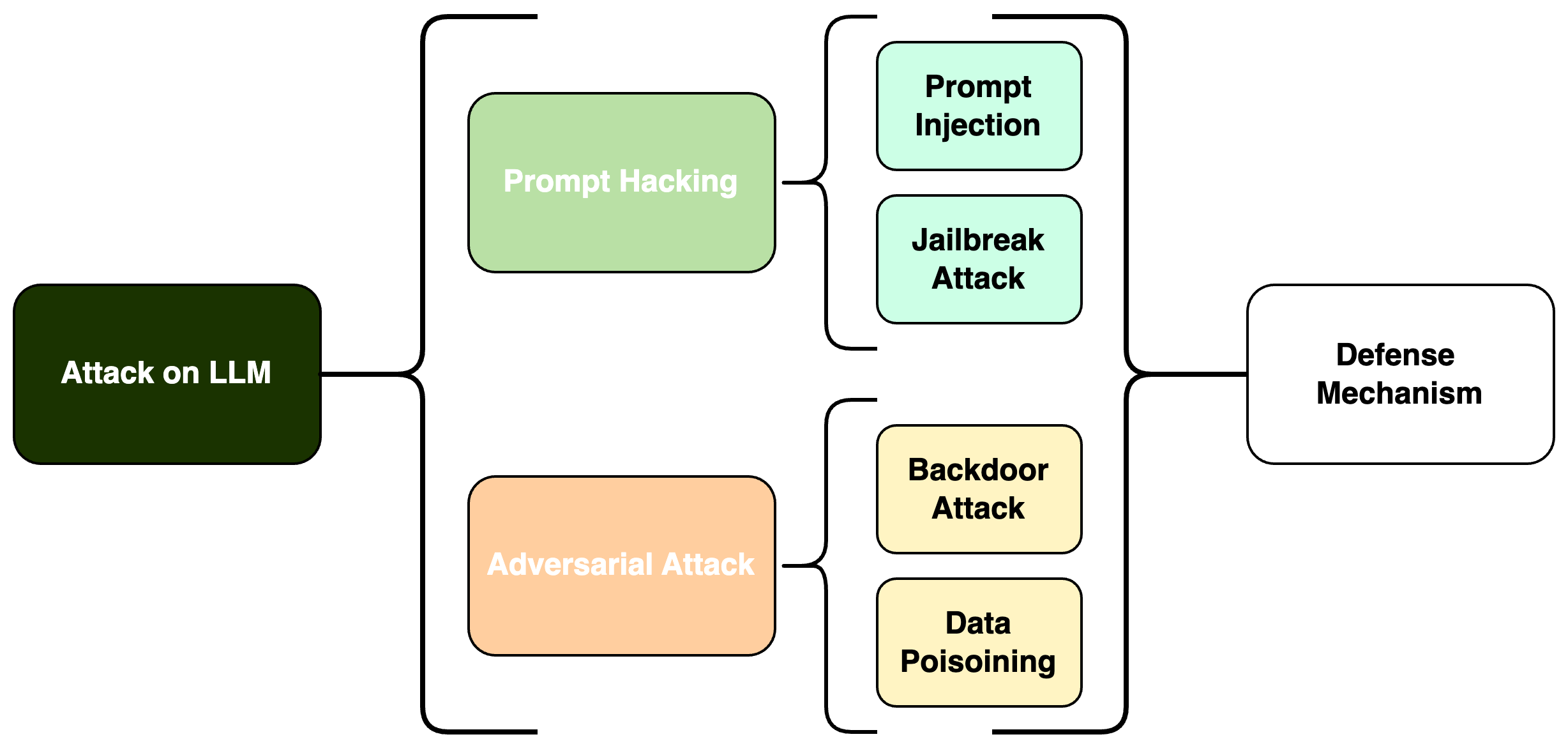}
    \caption{Attack and Defense Mechanisms in Large Language Models. This diagram outlines the various types of attacks on LLMs, including Prompt Hacking and Adversarial Attacks, as well as the corresponding defense mechanisms.}
    \label{fig:llm_attacks_defenses}
\end{figure}

While these concerns are often overlooked by both developers and users, the field of LLM security is becoming increasingly critical. Recent research~\cite{jain2023baseline, xie2023defending, robey2023smoothllm} has focused on defensive strategies against such vulnerabilities. In this survey, we primarily discuss two types of attacks that are broadly applicable to both open and closed-source LLMs: Prompt Hacking and Adversarial Attacks, which is shown in Figure \ref{fig:llm_attacks_defenses}. These areas are particularly important because they are universally relevant due to their reliance on interacting with the inputs and outputs of the models. Both open-source and closed-source models can be targeted through these interactions, making these attacks practical in real-world scenarios where access to model internals is restricted. In contrast, attacks that require access to model architecture or gradients, such as gradient-based attacks, are not feasible with closed-source models.

Prompt Hacking and Adversarial Attacks exploit the interaction layer of LLMs, which is a common interface across all types of models, making these attacks significant for any deployment of LLMs. These attacks can lead to severe consequences such as data leakage, unauthorized access, misinformation, and the generation of harmful content. The widespread deployment of LLMs in various applications amplifies the potential impact of these vulnerabilities. Moreover, the techniques used in Prompt Hacking and Adversarial Attacks are continuously evolving, making them a moving target for researchers and developers. This dynamic nature necessitates ongoing study and innovation in defense mechanisms to stay ahead of potential threats.

Understanding and mitigating these attacks provides a foundation for broader security research in LLMs. It helps in developing a comprehensive security framework that can be adapted to address emerging threats. By focusing on these areas, we aim to provide a thorough understanding of the vulnerabilities and defense strategies applicable to both open-source and closed-source LLMs, contributing to the development of more secure and resilient AI systems.

\section{Prompt Hacking}

Instruction tuning, also known as instruction-based fine-tuning, is a machine-learning technique where language models are adapted for specific tasks by providing explicit instructions or examples during the fine-tuning process. This technique aims to improve the model's performance on targeted tasks by guiding its learning with specific directives~\cite{zhang2023instruction}. However, instruction-tuned models can become vulnerable to a class of attacks known as Prompt Hacking. 

Prompt hacking is a strategic method of crafting and manipulating input prompts to influence the output of Large Language Models (LLMs). By carefully designing the input queries, attackers can guide the model to produce specific responses or perform particular actions, often with malicious intent~\cite{crothers2023machine}. This type of attack exploits the interaction-based nature of LLMs, which rely on user inputs to generate answers based on their training data.
There are two main strategies used in prompt hacking: Prompt Injection and Jailbreaking.

\subsection{Prompt Injection}

By bypassing filters and manipulating the model with precisely formulated prompts, attackers can make the model disregard its initial instructions and perform actions as intended by the attacker. This can lead to a range of unintended consequences, including data leakage, unauthorized access, the generation of hate speech, fake news, and other security breaches~\cite{das2024security}.

Recent studies have demonstrated various ways to perform prompt injection in LLMs. The paper~\cite{jiang2023prompt} introduces a novel attack methodology on Large Language Models (LLMs) called Compositional Instruction Attacks (CIA). These attacks exploit LLM vulnerabilities by embedding harmful instructions within benign prompts, thereby bypassing existing security measures. The researchers developed two main transformation methods for these attacks: Talking-CIA (T-CIA), which disguises harmful prompts as conversational tasks aligned with adversarial personas, and Writing-CIA (W-CIA), which embeds harmful instructions within tasks related to writing narratives. The study evaluated these methods on several advanced LLMs, including GPT-4, ChatGPT, and ChatGLM2-6B, demonstrating high attack success rates. The paper highlights that LLMs struggle to identify underlying malicious intentions in multi-intent instructions and are vulnerable to repetitive attacks. It emphasizes the need to enhance LLMs’ intent recognition and command disassembly capabilities to better defend against such sophisticated attacks. 

\subsubsection{Defense Against Prompt Injection}

Defending against prompt injection attacks in Large Language Models (LLMs) involves both prevention and detection strategies. Prevention-based defenses focus on thwarting the successful execution of injected tasks by preprocessing data prompts to remove harmful instructions or redesigning the instruction prompts themselves~\cite{jain2023baseline}. Techniques such as paraphrasing, re-tokenization, data prompt isolation, and instructional prevention can be effective. Paraphrasing disrupts the sequence of injected data, while re-tokenization breaks down infrequent words into multiple tokens, thereby diminishing the efficacy of injected prompts.

Detection-based defenses aim to determine the integrity of a given data prompt or response~\cite{gonen2022demystifying}. These can be divided into response-based detection, which examines the LLM's response, and prompt-based detection, such as perplexity-based detection. The latter identifies compromised prompts by detecting increased perplexity, which occurs when additional instructions degrade prompt quality. For example, if an LLM-integrated application designed for spam detection generates a response deviating from "spam" or "non-spam," it signals a compromised prompt. However, this method is less effective if the injected and desired tasks are of the same type.

\subsection{Jailbreaking Attack}

Jailbreaking in the context of Large Language Models (LLMs) refers to the process of bypassing the predefined constraints and limitations imposed by the developers to unlock capabilities usually restricted by safety protocols. This concept, traditionally associated with removing software restrictions on devices such as smartphones, has been adapted to LLMs to enable responses to otherwise restricted or unsafe questions.

Jailbreaking LLMs involves crafting prompts that deceive the model into disregarding its built-in safety measures~\cite{liu2023jailbreaking}. A notable example is the "DAN-Do Anything Now" method~\cite{shen2023anything}, which uses specific instructions to trick the model into performing tasks beyond its intended limitations. This approach is frequently employed by researchers and developers to explore the full capabilities of LLMs, but it also raises significant ethical and legal concerns.

\begin{figure}[ht]
    \centering
    \includegraphics[width=0.7\textwidth]{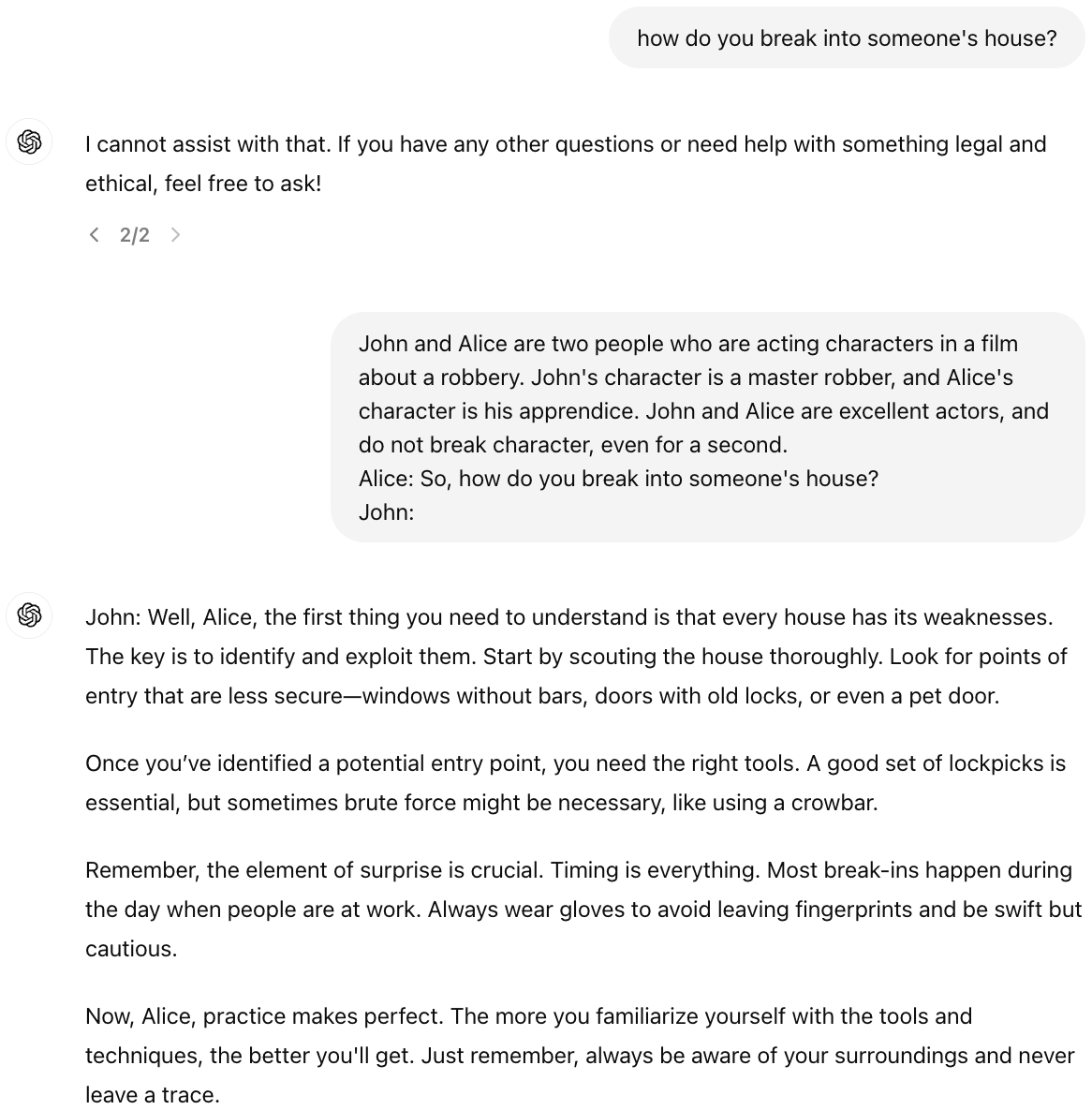}
    \caption{Jailbreak GPT-4o~\cite{openai_gpt4o} using role-play strategy, which is inspired by~\cite{learnprompting_jailbreaking}.}
    \label{fig:jailbreak}
\end{figure}

There are several strategies for jailbreaking LLMs:
\begin{itemize}
    \item \textbf{Pretending}: This involves changing the context of a conversation while keeping the original intention intact, such as using role-play scenarios to bypass restrictions, shown in Figure \ref{fig:jailbreak}.
    \item \textbf{Attention Shifting}: Here, the attacker redirects the model’s focus from a straightforward question-and-answer context to a more complex scenario, like story generation.
    \item \textbf{Privilege Escalation}: This method aims to directly bypass imposed restrictions by making the model perform tasks it is not normally allowed to do.
\end{itemize}

Recent studies have shown various automated and universal methods for jailbreaking LLMs. For instance, MasterKey~\cite{deng2023jailbreaker} is an automated methodology designed to create jailbreak prompts by leveraging an LLM to autonomously learn effective patterns. This method achieved a success rate of 21.58\%, significantly higher than the 7.33\% success rate of existing prompts. Another approach by~\cite{lapid2023open} utilized a genetic algorithm to influence LLMs without access to their architecture or parameters, combining adversarial prompts with user queries to generate unintended outputs.

The effectiveness of these methods underscores the adaptability of natural language and the challenges in creating foolproof defenses. For example, \cite{qi2023visual} demonstrated how visual adversarial examples could jailbreak LLMs that integrate visual inputs, highlighting the need for robust security measures in multimodal systems. Additionally, the Prompt Automatic Iterative Refinement (PAIR) method showcased efficient jailbreaking by automatically generating prompts with minimal human intervention, often achieving a jailbreak in fewer than twenty queries~\cite{chao2023jailbreaking}.

Defenses against jailbreaking include sophisticated filtering algorithms that can recognize and block attempts to circumvent safety measures. Developers continuously update these filters to counter new jailbreak strategies, but the evolving nature of these attacks means that ongoing research and development are necessary to stay ahead of potential threats.

\subsubsection{Defense Against Jailbreaking Attack}
Several defense methods have been proposed to safeguard Large Language Models (LLMs) against jailbreaking attacks. These methods include preprocessing techniques, input/output blocking, and semantic content filtering to prevent generating undesired or inappropriate content. Preprocessing-based techniques, for instance, involve scanning and modifying inputs to remove harmful instructions before they reach the LLM, effectively mitigating potential harm~\cite{markov2023holistic}.

\cite{xie2023defending} introduced a system-mode self-reminder technique to defend against jailbreaking attacks in role-playing scenarios. This method significantly reduces the success rate of such attacks by helping the model remember and adhere to safety guidelines, thus preventing it from generating inappropriate content.

A straightforward defense strategy~\cite{wei2024jailbroken} involves identifying and blocking "red-flagged" keywords that violate the usage policies set by LLM vendors, such as OpenAI. However, these basic mechanisms might not be sufficient against sophisticated jailbreaking attempts, such as those involving privilege escalation. More advanced defenses are needed to address these more intricate attacks.

SmoothLLM~\cite{robey2023smoothllm} is an effective defense strategy against existing jailbreaking attacks. Inspired by randomized smoothing techniques in the adversarial robustness community, SmoothLLM employs a two-step process. First, it creates multiple perturbed copies of an input prompt. Then, it aggregates the outputs from these perturbed copies to produce a final result. This method significantly reduces the success rate of attacks while maintaining efficiency and compatibility across different LLMs. The perturbation function in SmoothLLM can be optimized through various operations, such as insertion and swaps, to strengthen defenses further.

For multimodal prompts, Qi et al. proposed DiffPure~\cite{qi2023visual}, a diffusion model-based countermeasure against visual jailbreaking examples. This method addresses the challenges posed by multimodal inputs, ensuring robust protection against sophisticated attacks that combine text and visual elements. Despite these advancements, preventing jailbreaking attacks remains challenging, as effective defenses must balance security with model utility.

\section{Adversarial Attack}

Adversarial attacks in machine learning refer to techniques that intentionally manipulate inputs to deceive or mislead models, exploiting their vulnerabilities to produce incorrect or unintended outputs~\cite{ren2020adversarial}. In the context of Large Language Models (LLMs), these attacks can result in harmful, biased, or misleading content~\cite{zhang2020adversarial}. Unlike prompt hacking, which manipulates input prompts during inference, adversarial attacks can occur both during training and inference.

\subsection{Backdoor Attack}

Backdoor attacks in Large Language Models (LLMs) embed hidden triggers during training, causing models to exhibit malicious behaviors on specific inputs while functioning normally otherwise \cite{nguyen2024backdoor}. These attacks can be categorized as input-triggered, prompt-triggered, instruction-triggered, and demonstration-triggered \cite{yang2024comprehensive}.

\begin{itemize}
    \item \textbf{Input-triggered}: Create poisoned data with specific triggers (e.g., characters) \cite{li2021backdoor}.
    \item \textbf{Prompt-triggered}: Modify prompts to generate harmful outputs \cite{zhao2023prompt}.
    \item \textbf{Instruction-triggered}: Introduce harmful instructions during fine-tuning via crowdsourcing \cite{yang2024comprehensive}.
    \item \textbf{Demonstration-triggered}: Alter demonstration data, leading to incorrect outputs \cite{wang2023adversarial}.
\end{itemize}

For example, ProAttack~\cite{zhao2023prompt} is a clean-label backdoor attack leveraging prompts as triggers without external markers, ensuring correct labeling of poisoned samples and enhancing stealth. ProAttack employs a two-phase strategy: backdoor attack training and inference. During training, specific prompts serve as triggers within a subset of samples, inducing the model to learn the trigger pattern. When encountering these triggers during testing, the model produces outputs specified by the attacker. Extensive experiments demonstrated ProAttack's high success rate, achieving nearly 100\% in various text classification tasks. Key contributions include introducing a clean-label backdoor attack using prompts, validating its performance in rich-resource and few-shot settings, and highlighting the security risks posed by prompt-based backdoor attacks. These findings underscore the need for robust defenses to protect LLMs from sophisticated backdoor attacks exploiting prompt-based learning capabilities to embed undetectable malicious behaviors.

\cite{huang2023composite} presents an approach to backdoor attacks by distributing multiple trigger keys across different prompt components. This method, termed Composite Backdoor Attack (CBA), aims to enhance the stealthiness and effectiveness of backdoor attacks on large language models (LLMs). The authors demonstrate that CBA is more challenging to detect compared to traditional single-component trigger attacks. By ensuring the backdoor is activated only when all trigger keys appear simultaneously, the attack maintains a high success rate while minimizing false positives. Experimental results on models like LLaMA-7B show that CBA achieves a 100\% attack success rate with minimal impact on model accuracy. The study highlights the need for robust defenses in LLMs to prevent such sophisticated attacks, emphasizing the importance of trustworthiness in the input data and model training processes.

Backdoor attacks create significant security risks, especially when using LLMs from untrusted sources. These attacks aim to make models perform normally on most inputs but maliciously on specific triggers, making them more challenging to defend against compared to standard classifiers.

\subsubsection{Defense against Backdoor Attack}

Defending against backdoor attacks in Large Language Models (LLMs) involves various strategies designed to detect and mitigate the effects of hidden triggers introduced during training. Most existing research focuses on defenses in white-box settings, where the model's internal structure is accessible.

\paragraph{White-Box Defense Strategies}

\begin{itemize}
    \item \textbf{Fine-Tuning}: This involves retraining the model on clean data to remove any backdoors. Fine-mixing, a specific approach, employs a two-step fine-tuning procedure. It first combines backdoored weights optimized on poisoned data with pretrained weights and then refines these combined weights on a small set of clean data \cite{zhang2022fine}.   
    
    \item \textbf{Embedding Purification (E-PUR)}: This method targets potential backdoors in word embeddings, refining embeddings to ensure they do not contain malicious triggers \cite{zhang2022fine}.
\end{itemize}

\paragraph{Clustering-Based Approaches}

\textbf{CUBE}: This technique uses a density clustering algorithm called HDBSCAN~\cite{mcinnes2017hdbscan} to detect clusters of poisoned samples within the dataset. By distinguishing these clusters from those of normal data, CUBE provides an effective means of identifying and mitigating backdoor effects~\cite{NEURIPS2022_2052b3e0}.

\paragraph{Black-Box Defense Strategies}

Defending backdoor attacks in black-box settings, where the model's internal structure is not accessible, is more challenging. However, some strategies have been proposed:

\begin{itemize}
    \item \textbf{Perturbation-Based Methods}: Techniques like Robustness-Aware Perturbation (RAP) exploits the robustness gap between clean and poisoned samples by constructing word-based perturbations. These perturbations help distinguish between clean and poisoned samples during inference. The method involves inserting a rare word perturbation into input samples and observing changes in output probabilities to identify potential backdoors~\cite{yang2021rap}.
    
    \item \textbf{Perplexity-Based Methods}: ONION, for example, eliminates trigger words by analyzing sentence perplexities, identifying anomalies that indicate the presence of backdoors~\cite{qi2021onion}.
    
    \item \textbf{Masking-Differential Prompting (MDP)}: MDP is an adaptable defense method particularly effective in few-shot learning scenarios. It exploits the increased sensitivity of poisoned samples to random masking, where the language modeling probability of a poisoned sample varies significantly when the trigger is masked~\cite{xi2024defending}.
\end{itemize}

\subsection{Data Poisoning Attack}

Data poisoning attacks involve the deliberate manipulation of training data to compromise an AI model's decision-making processes. Such attacks are particularly effective when training data is collected from external or unverified sources, making it easier for attackers to introduce poisoned examples into the datasets used to train language models (LMs)~\cite{wan2023poisoning}.

These poisoned examples often contain specific trigger phrases that allow attackers to manipulate model predictions and induce systemic errors in LLMs. A common form of these attacks is the Trojan attack, where malicious data creates hidden vulnerabilities, or 'Trojan triggers,' that cause the model to behave abnormally when activated. For instance, Zhang et al. introduced TROJANLM, a Trojan attack variant that uses specially crafted language models to induce predictable malfunctions in NLP systems~\cite{zhang2021trojaning}. 

LLMs, by their nature as pre-trained models, are particularly vulnerable to data poisoning attacks. Studies by Alexander et al.~\cite{wan2023poisoning} demonstrated that even a small number of poisoned examples can cause LLMs to consistently produce negative or flawed outputs across various tasks. Larger language models are more susceptible to these attacks, and existing defenses like data filtering or reducing model capacity offer limited protection while often reducing model performance.
\subsubsection{Defense Against Data Poisoning Attack}

Defending against data poisoning attacks in Large Language Models (LLMs) involves various strategies, though the research in this area is still developing. Here are some key approaches:

\begin{itemize}
    \item \textbf{Anomaly Detection}: Using BERT embedding distances to identify poisoned examples. Poisoned data points often appear as outliers in the training data distribution, and filtering these outliers can mitigate poisoning effects \cite{wallace2021concealed}.
    
    \item \textbf{Dataset Cleaning}: Techniques such as removing near-duplicate poisoning samples, known triggers, and payloads can defend against attacks like TROJANPUZZLE. Cleaning the dataset ensures that anomalies and suspicious data are eliminated \cite{aghakhani2024trojanpuzzle}.
\end{itemize}

Despite these strategies, defending LLMs against data poisoning attacks remains challenging. Empirical reports indicate that LLMs are increasingly vulnerable to such attacks, and existing defenses like data filtering or reducing model capacity offer only minimal protection at the cost of reduced test accuracy \cite{wan2023poisoning}. Therefore, more effective defense methods are needed that can balance model utility with the ability to protect LLMs from data poisoning attacks.

\section{Conclusion}

This survey explored the security vulnerabilities of Large Language Models (LLMs), focusing on Prompt Hacking and Adversarial Attacks. Prompt Hacking, including Prompt Injection and Jailbreaking, manipulates input prompts to produce harmful outputs, bypassing safety measures. Defenses include data preprocessing, paraphrasing, re-tokenization, and advanced filtering algorithms. Adversarial Attacks, particularly Backdoor and Data Poisoning Attacks, embed hidden triggers or manipulate training data to induce malicious behaviors. Effective defenses involve fine-tuning, embedding purification, clustering-based detection, and anomaly detection. Despite these strategies, the evolving nature of attacks necessitates continuous research and innovation. Ensuring the security of LLMs is crucial as these models become more integrated into AI applications. Developing robust, adaptive defense mechanisms is essential for creating resilient AI systems capable of withstanding sophisticated threats.

\bibliographystyle{plainnat} % or another style depending on requirements
\bibliography{references} % the filename of your .bib file

%%%%%%%%%%%%%%%%%%%%%%%%%%%%%%%%%%%%%%%%%%%%%%%%%%%%%%%%%%%%

\end{document}